# A Novel Psychometrics-Based Approach to Developing Professional Competency Benchmark for Large Language Models

Elena Kardanova, Alina Ivanova, Ksenia Tarasova, Taras Pashchenko, Aleksei Tikhoniuk, Elen Yusupova, Anatoly Kasprzhak, Yaroslav Kuzminov, Ekaterina Kruchinskaia, Irina Brun

National Research University Higher School of Economics, Moscow, Russia

**Abstract**

The era of large language models (LLM) raises questions not only about how to train models, but also about how to evaluate them. Despite numerous existing benchmarks, insufficient attention is often given to creating assessments that test LLMs in a valid and reliable manner. To address this challenge, we accommodate the Evidence-Centered Design (ECD) methodology and propose a comprehensive approach to benchmark development based on rigorous psychometric principles. In this paper, we have made the first attempt to illustrate this approach by creating a new benchmark in the field of pedagogy and education, highlighting the limitations of existing benchmark development approach and taking into account the development of LLMs. We conclude that a new approach to benchmarking is required to match the growing complexity of AI applications in the educational context.

We construct a novel benchmark guided by the Bloom's taxonomy and rigorously designed by a consortium of education experts trained in test development. Thus the current benchmark provides an academically robust and practical assessment tool tailored for LLMs, rather than human participants. Tested empirically on the GPT model in the Russian language, it evaluates model performance across varied task complexities, revealing critical gaps in current LLM capabilities. Our results indicate that while generative AI tools hold significant promise for education—potentially supporting tasks such as personalized tutoring, real-time feedback, and multilingual learning—their reliability as autonomous teachers' assistants right now remain rather limited, particularly in tasks requiring deeper cognitive engagement.

## 1 Introduction

In evaluating the performance of Large Language Models (LLMs), benchmarks play a crucial role, representing standardized sets of tasks with established evaluation criteria (Guo et al., 2023; Naveed et al, 2024). Despite the diversity of existing benchmarks covering a wide range of cognitive abilities of LLMs, from solving mathematical problems to logical reasoning, they face some limitations. Among the most discussed issues are benchmark obsolescence (McIntosh et al., 2024), the problem of "data leakage" (Zhu et al., 2024) or "data contamination" (Roberts, 2023). However, there are also challenges in interpreting results of LLM evaluation using benchmarks from the perspective of LLM professional competencies. This is due to the poor focus of most existing benchmarks on the taxonomy of cognitive abilities and professional outcomes, resulting in a lack of benchmark specifications linking model metrics, content blocks, levels of cognitive complexity according to the taxonomy, and professional outcomes.

This work offers a new psychometrics-based framework of developing benchmarks for evaluating LLM professional competencies. The novelty of the approach lies in an attempt to apply (with some modifications that we further consider in section three) rigorous principles of human test development towards benchmarks development. We illustrate this approach with a step-by-step description of our benchmark development process, beginning with the formulation of educational outcomes and blueprint creation, and progressing to the development of multiple tasks based on varying taxonomies and difficulty levels, crafted by specially trained professional experts. This approach will not only resolve the issue of data leakage and contamination but also enable a meaningful interpretation of LLM's test results, a crucial step toward a more precise and unbiased assessment of their cognitive abilities.

The article unfolds across five main sections: first, we outline existing research on LLM benchmarks, including discussions on LLM competencies and the limitations of current evaluation methods; second, we introduce the study's scope, objectives, and unique methodological approach; third, we thoroughly describe the process of the benchmark in Pedagogy and Education development, aligning it with psychometric test development practices; fourth, we present the outcomes from empirical testing of our benchmark with GPT model and offer some interpretations of the results; and finally, we summarize the contributions of the study and propose some future research directions.

*Global Practices in Benchmarking LLMs*

LLMs are at the forefront of contemporary AI technologies, with computational capabilities and architectures in constant evolution. Evaluating LLM performance is a rapidly growing area. Benchmarks for LLM evaluation offer several advantages, allowing for the evaluation of model speed, accuracy, and efficiency. There is a vast array of different benchmarks, most of which are publicly available on services such as HuggingFace. Benchmarks for LLMs can be grouped into subgroups: benchmarks for assessing natural language understanding (NLU) and natural language generation (NLG), benchmarks for knowledge and reasoning skills[1] encompassing areas as diverse as legal studies (Fei et al., 2023) and radiation oncology (Huang et al., 2023), benchmarks for holistic assessment (Guo et al., 2023), and ethical benchmarks (Lourie et al., 2021) and others. Often these benchmarks consist of multiple-choice questions.

Probably one of the most used benchmarks—the MMLU benchmark devised by Hendrycks et al. (2020)—comprises 15,908 multiple-choice questions across 57 distinct subjects, ranging from high school algebra to world history. MMLU functions as an indicative measure of LLMs' overarching world knowledge, a reflection of their general intelligence. The newly released MMLU-Pro is said to extend the original MMLU benchmark by integrating more difficult reasoning-focused questions and expanding the choice set from four to ten options. Additionally, MMLU-Pro eliminates the trivial and noisy questions in MMLU, resulting in 12,032 questions in total and posing many more challenges to LLMs than the previous version (Wang et al., 2024).

However, researchers express a certain level of doubt about whether the most used benchmarks measure what they are supposed to in a valid and reliable way (Fang, Oberski, & Nguyen, 2024). The question is not idle, as the main goal of benchmarks is to evaluate the performance of models, to compare different models with each other (Zhou et al., 2023), and, as a result, to support the precise selection of an LLM best suited to a given purpose. However, this approach also restricts the depth of interpretation regarding the LLM's domain-specific expertise for several reasons.

Firstly, there are studies that suggest LLMs can acquire "traits" or "skills" (speaking figuratively) from the rich texts on which they are trained (e.g., Zhuang et al., 2023; Pellert et al., 2024). LLM researchers argue about the need in developing items based on principles for psychological and educational tests' development (Pellert et al., 2024; Fang et al., 2024). These principles for assessing cognitive, non-cognitive and professional skills require a theoretical framework that defines and describes the content and its complexity, the scope and the levels of the construct under investigation. Many benchmarks are created by crowd-sourcing human writers (e.g., GSM8K, Cobbe et al., 2021) or collected from convenient sources. The above-mentioned MMLU dataset (Hendryks et al., 2020) comprises questions that have been handpicked from various online sources that are openly accessible, such as the Graduate Record Examination (GRE) materials or practice questions for the Examination for Professional Practice in Psychology. As a result, benchmarks often include many tasks, but it is unclear whether they assess a specific construct or proficiency in a specific field (Wang et al., 2023b).

---

[1] According to the latest release of Open AI the company "developed a new series of AI models designed to spend more time thinking before they respond. They can reason through complex tasks and solve harder problems than previous models in science, coding, and math". Retrieved 24.09.2024 from URL: https://openai.com/index/introducing-openai-o1-preview/

Secondly, there are few studies that consider the psychometric properties of benchmark tasks, such as the difficulty of the tasks or their discriminative power (Fang et al., 2024). Proper psychometric analysis of items used for LLM assessment is important to draw inferences about the level of LLMs proficiency and reliably differentiate models according to their level. Even though reliability is claimed to be a principle of benchmark development (Xiao et al., 2023), this term is considered more broadly in psychometrics than in computer science. In computer science the reliability of a benchmark is understood as "replicability" or "robustness", and in psychometrics test reliability allows us to identify inconsistencies in results between different test elements, raters and data subsets (Wang et al., 2023b).

Thirdly, according to Wang and colleagues (2023b), many benchmarks are task-oriented and lack predictive or exploratory power. The lack of predictive power does not allow us to predict whether an LLM will solve new unforeseen tasks. Lack of exploratory power imposes restrictions on explaining why LLMs solve some tasks and fail the others.

Finally, researchers often compare LLM' test results with those of human examinees (e.g., Dao et al., 2023). But there is still a controversial point regarding such a comparison. The main argument against comparing performance on tests and professional exams between LLMs and people is that most tests and professional exams test subject matter knowledge and underestimate real-world skills (Narayanan & Kapoor, 2023). In other words, these tests and exams overestimate exactly what language models are good at. At the same time, there are several studies that use item response theory methods to align LLMs test results with human test results (Fang et al., 2024). However, even the authors of such studies make such alignments

As mentioned in (Wang, et al, 2023b) traditional benchmarks allow evaluate LLMs only in terms of tasks included in them—for example, translation, summarization, sentiment analysis, etc. This approach to benchmark development and LLM evaluation is not appropriate if we want to evaluate LLM's professional competencies, for example, in law, education, etc. For instance, although a benchmark contains tasks related to education and pedagogy, it is insufficient to evaluate an LLM acting as a general teacher assistant due to the variety of potential situations that could arise. The goal of the current paper is to suggest using psychometrics in a broad sense, including the principles of test development, as a comprehensive theoretical approach for the benchmark development as an alternative to the most commonly used approach in the LLM evaluation community.

*The novelty of this research*

Psychometrics offers a construct-oriented approach to developing tests (Wang et al, 2023b). Various constructs are thought to be latent variables that underlie certain behavior and can predict it in different contexts, regardless of what tasks are used for it. In other words, latent constructs can be the cause of performance patterns across a variety of tasks, and psychometrics can predict performance across a wide range of tasks and situations.

The idea to use psychometrics for benchmarks development and LLM evaluation is not novel, but still underinvestigated. Recently, researchers proposed PATCH: a framework for psychometrics-assisted benchmarking of LLMs (Fang et al., 2024), consisting of 8 steps from defining the construct to LLM proficiency score production. The implementation of PATCH was piloted with GPT-4 and Gemini in 8th grade mathematics using the released test items and real student's data from Trends in International Mathematics and Science Study (TIMSS) of 2011. In their study Fang and colleagues (2024) demonstrate that their approach yields results that diverge from traditional benchmarking practices, offering a more comprehensive and informative perspective.

In another relevant work, Dao and colleagues (Dao et al., 2023) proposed a dataset generated from the Vietnamese National High School Graduation Examination and comparable tests that covers nine subjects. The research team included in the dataset 300 literary essays and over 19,000 questions, allowing assessment of LLMs in multitasking situations such as question answering, text generation, reading comprehension, visual question answering, and more. The performance of ChatGPT and BingChat is contrasted with that of Vietnamese students. Interestingly, the authors base the development of their dataset on Bloom's taxonomy, which makes the benchmark development process more similar to test development practices. The dataset has items with four cognitive levels—knowledge (easy items), comprehension (intermediate items), application (difficult items), and high application (extremely tough). As such, the study allows for a more nuanced analysis of difficulty, what enhances the capacity for in-depth analysis.

It's important to note that currently many benchmark developers use existing sets of items, which may have been used for the pre-training of LLMs. Additionally, these tasks were initially developed for human beings, who usually do not have immediate access to limitless databases, get tired, experience emotions and have other features that call for special requirements to the tasks they should be assessed with (Haladyna & Rodriguez, 2013). Nevertheless, the scholarly endeavors undertaken by the aforementioned researchers have served as a catalyst for the advancement of our own work, which seeks to contribute to the field through the development of a psychometrically robust assessment specifically tailored for LLMs.

Our approach surpasses previous methodologies by incorporating the following innovative strategies:

1) We have constructed a novel benchmark dedicated to the evaluation of LLMs, distinct from any pre-existing examination or testing corpus. This dataset is a pioneering tool, purpose-built for the assessment of LLMs' proficiency, inspired by the principles of modern psychometrics.

2) This benchmark has been meticulously crafted by a consortium of experts in the field of pedagogy and education, each of whom was trained in fundamental test development principles. Their collective expertise has ensured the creation of a dataset that epitomizes academic rigor and practical relevance.

3) The assessment we have devised is intended explicitly for the evaluation of LLMs, rather than human participants. This denotes a paradigm shift in assessment development, with a clear focus on the unique capabilities and requirements of artificial intelligence systems.

4) We have designed our assessment to gauge the performance of the LLM when interacting in the Russian language. The development of non-English language benchmarks represents a necessary step in broadening the scope of LLM testing and ensuring its applicability across diverse linguistic contexts.

# 2 Benchmark development process

## 2.1 Differences in development process for humans and LLMs

If one compares tests for humans and LLM-targeted benchmarks, the most apparent difference is their size. The latter often have thousands of tasks (i.e., Hendrycks et al., 2020), while the former rarely exceed a dozen items. Tasks for LLMs may contain up to 10-20 distractors, while a typical multiple-choice question intended for human examinees contains 4-7 answer options. The limitations of human attention and focus necessitate a radically different approach to test development compared to benchmark development.

From a psychometrics point of view, test development emphasizes a clear delineation of the target construct and expert-driven item generation. This ensures decent sampling of target behavior and a clear link between the behavior the task must elicit and the target construct. Psychometrically grounded tests focus on item creation and refinement, whereas LLM benchmarks focus more on dataset compilation and quality control (Davis, 2016; Narayanan & Kapoor, 2023).

Psychometrics-based tests require pilot studies and detailed data analysis (within classical test theory or item response theory approaches) with subsequent changes before the actual use by the audience. The International Test Commission (ITC) Guidelines on Quality Control in Scoring, Test Analysis and Reporting of Test Scores require for any test form reporting basic statistics about item and test functioning and guidelines for interpreting the test scores (ITC, 2012). LLM benchmarks focus more on error and agreement analysis and less on detailed statistical validation (Fang et al., 2024).

Logic of the test and benchmark development process is presented in Table 1.

*Table 1*
*Comparison of test development process (based on Fang et al., 2024)*

| Development process | Benchmark | Test |
|---|---|---|
| Initial stages | Test need (& construct) specification<br>Overall planning | Construct and test need specification<br>Overall planning |
| Item development | Item collection<br>       Quality control<br>OR<br>Item creation and/or | Item generation and review<br>Piloting items<br>Item quality analysis |

|  | annotation<br>Item generation<br>Agreement analysis<br>Error analysis |  |
|---|---|---|
| Test construction | Dataset construction | Test construction and specification |
| Test functioning study | Model selection and evaluation | Psychometric quality analysis<br>Test scoring and norming |
| Release | Benchmark release | Technical manual<br>Test release |

The differences in the described processes stem from two radically different answers to the question: How can we ensure that the items are measuring what we want them to measure and, being combined in a test, give us a clear indication of the target construct? Test development approach is based on the causal theory of measurement (Markus & Borsboom, 2013), selecting tasks to eliminate rival explanations of elicited behavior and sampling them to cover all aspects of the target construct. A benchmarking approach follows a representational logic (Michell, 2007), forming a test that is a representative sample of behavior that is elicited by the underlying quality. The representational logic has the potential to make test results very generalizable. Traditional tests, featuring few standardized items stripped of any sources of error relevant for real-life performance of LLMs, offer less generalizability.

However, benchmarks often fall into other pitfalls. A lack of clear operational construct definition makes it impossible to demonstrate that the test adequately represents the construct. Moreover, adequate representation of the construct has a qualitative dimension, as a total score reflects the construct aspects in proportion to the presence of items loading on these constructs. A smaller number of items in psychometrics-informed test development allows for a much deeper analysis of their respective constructs.

However, one must not forget that the underlying mechanisms of how LLMs and humans respond to an item are not alike. While test outcomes with human subjects allow for interpretations that rely on psychological, cognitive, and neurobiological concepts, LLM behaviors are simply predictions of the next most likely token (Hagendorff, 2023). This is why the test development approach cannot be adapted to LLM assessment without significant changes.

An optimal assessment instrument for LLMs can be developed by capitalizing on the strengths of both - a test-development approach and a benchmarking approach. It is necessary to evaluate both the generation and understanding capabilities of LLMs and make inferences in a way dissimilar to reasoning about humans, for whom understanding comes much easier. Moreover, input characteristics—such as complexity, clarity, length, and the number of instructions—are significantly more important for LLMs than for humans, and should be given careful consideration.

2.2 The psychometric approach to benchmark development

Recent research in the field of LLMs has shown a particular interest in evaluating the extent to which different LLMs can assist in assessing knowledge at critical learning checkpoints (e.g., during professional examinations or certification processes). Such studies have already been conducted in various domains, including medicine (Gilson, Safranek, Huang, Socrates, Chi, Taylor, Chartash, 2022; Wang, Gong, Jia, Xu, Zhao, ..., Li, 2023a), law (Fei, Shen, Zhu, Zhou, Han, Zhang, ..., Ge, 2023), and chemistry (Guo, Nan, Liang, Guo, Chawla, Wiest, Zhang, 2023), among others. In the introduction we already mentioned why this is insufficient to evaluate LLM's performance in a target domain.

In this paper we adopt the psychometrics approach to the benchmark development for LLM. Psychometrics offers a unique approach to test development—construct-oriented approach (Wang, et al, 2023b). This approach is based on the causal theory of measurement, which suggests that item responses are caused by the property being measured, and, for example, the number of items is not important itself, but the proof that responses to items are not caused by anything other than the target characteristic is crucial. In this case, the number of items may be smaller, but their quality is important, they must relate to all structural components of the construct being measured, and all alternative explanations for responses to items must be eliminated. Psychometrics provides a mechanism for how to develop tests from the operationalization of the construct to the analysis of the test results.

In this work, we also draw upon the Evidence-Centered Design (ECD) framework used in educational assessment that has been developed in psychometrics (Mislevy & Haertel, 2006; Zieky, 2013, Oliveri & Mislevy, 2019). This theoretical approach allows one to identify observable evidence of construct's manifestation and thereby move from a general definition of the construct to the specific variables on which test items are created. In particular, this method allows for the creation of items that use real-life situations as a context, which allows to assess skills in situations that approximate real-life situations (rather than just assessing knowledge of theoretical concepts).

Research in this area is ongoing, and a recent development introduced the Evidence-Centered Benchmark Design (Liu et al., 2024), an ECD-based framework that systematically organizes the benchmark design process and requires benchmark designers to define, justify, and substantiate their design decisions—such as explicitly identifying the target construct and detailing the methods for collecting evidence from model responses regarding those capabilities. Thus, ECD conceptualizes benchmark creation as a process of collecting evidence regarding LLM competencies and offers guidance on the development, while in our case, enabling a construct-oriented approach. We have reinterpreted the core models of the ECD conceptual assessment framework — Proficiency Model, Task Model, Evidence Model—with a particular focus on a deeper understanding of the Task Model.

*The Proficiency Model*

The Proficiency Model delineates the characteristics or competencies that the assessment aims to evaluate. Its purpose is to establish a clear linkage between the benchmark and its intended objective. To accomplish this, benchmark developers must define the professional competencies of LLM, substantiate the asserted connection, and provide a robust rationale for the conceptualization of these competencies. When applying to the task of evaluation of LLM professional competencies, it means we should formulate the educational outcomes – what we would like to assess. This is done through the process of operationalization, that is the first step in the process of any test development, including benchmark for LLM evaluation. It involves formulating educational outcomes in terms of measurable indicators to describe the substantive area accurately and unambiguously. This ensures that the selected/developed tasks allow for valid conclusions about the model's knowledge/skills/abilities in that substantive area. The educational outcomes should be action-oriented, observable, and clearly articulated.

*The Task Model*

The Task Model facilitates the verification that the developed benchmark is indeed suitable for measuring the competencies of LLM as outlined in the Proficiency Model. Benchmark developers must justify, through the characteristics of the test items, how each item elicits evidence related to the targeted LLM competencies. Thus, the test content must be defined. It can be organized as a structured list of test content units that belong to each sub-domain of professional field. This step is necessary to ensure that all elements of the subject area content required for testing are included in the benchmark. It is important to note that the content specification should align not with the corpus of materials on which the LLM was trained, but rather with the standard of knowledge and skills that the LLM needs to achieve in order to be an effective assistant in the professional field assessed. If the selection of content for benchmark tasks is tailored to the training corpus of the neural network or if there is a deliberate effort to create tasks that the LLM cannot answer, the tasks will cease to reflect the aspects of knowledge and professional activity that are important in human work (and in which the neural network should assist). Instead, they will be adjusted to fit the results. Consequently, the LLM may exhibit high performance that is not useful to humans or that quickly becomes obsolete.

Two aspects are crucial when we think on how to evaluate professional competencies of LLM:

1) The breadth of the LLM's "knowledge"

2) The depth of the LLM's "knowledge"

The breadth of knowledge possessed by LLMs is contingent upon the data on which the neural network was trained, as well as the quantity and nature of documents available to it for specific content units (e.g., the birthdate of a historical figure). LLMs "store" the knowledge acquired during training within their parameters (Roberts, Raffel, Shazeer, 2020). In their study, Kandpal, Deng, Roberts, Wallace, and Raffel (2023) investigated the relationship between the information learned by a LLM and the amount of information related to specific content elements present in the pre-training dataset. In other words, they examined how the frequency of information in the training data affects the model's ability to learn and accurately handle that information. Their findings indicate that LLMs perform well in open-domain QA benchmarks only when the content element frequently appears in the training dataset. Conversely, if an element is infrequently encountered, LLMs struggle to "learn" and accurately manipulate that content element (Kandpal, Deng, Roberts, Wallace, Raffel, 2023, p. 8). Additionally, several studies have demonstrated that LLMs can "recall" factual knowledge with high accuracy even without fine-tuning (e.g., Petroni, Rocktäschel, Lewis, Bakhtin, Wu, Miller, Riedel, 2019).

The second aspect of LLM knowledge, their depth, is much less explored. This may be due to the inherent difficulty in assessing the depth of knowledge (whether in humans or neural networks), which necessitates the use of an additional tool—a model that classifies the depth of comprehension. Such models are known as taxonomies. Their purpose is to align desired educational outcomes (educational goals), as stipulated in various documents (e.g., Federal State Educational Standards), with actual learning outcomes and their measurement.

Numerous taxonomies of educational objectives exist. The most popular among them is Bloom's Taxonomy (Bloom, Engelhart, Furst, Hill, Krathwohl, 1956) and its revised version (Krathwohl, 2002). However, there are dozens, if not hundreds, of other taxonomies: subject-specific and independent, ordered by cognitive complexity and not, designed for specific measurement tools and educational programs. Examples include the SOLO Taxonomy (Collis, Biggs, 1986), Wilson's Taxonomy (Nayef, Yaacob, Ismail, 2013), and Webb's Taxonomy (Webb, 2002). This diversity presents a challenge: any of the existing taxonomies could be used to evaluate LLMs, and it is necessary to determine which one would be most suitable.

However, conducted studies (Kassner, Krojer, Schütze, 2020) indicate that even well-trained models are unable to make deep inferences from incomplete data. In other words, the higher levels of taxonomy, which engage higher cognitive functions (analysis, synthesis etc.), are likely to remain inaccessible to the model. Thus, the model will consistently perform worse on tasks and in domains where knowledge is less systematized, contains numerous contradictions, and exceptions to rules. Thus, choice of the taxonomy is the third step in the development of benchmarks for adequately assessing the model's knowledge and skills.

Modeling the internal connections between cognitive abilities and explaining how an LLM links various 'pieces' of information enables more accurate evaluation of its results. This, in turn, allows the benchmark development process to be grounded in cognitive measurement theory, reinforcing and complementing the representativist approach. For example, the researchers (Fei et al., 2023) suggest for their benchmark a modified version of Bloom's taxonomy consisting of three levels: 1) reproduction, 2) understanding, 3) application. Such a breakdown by depth enables us to obtain more precise information about what exactly the LLM does and does not know, and in what types of tasks it makes incorrect predictions.

Thus, the taxonomy allows to determine the depth of a LLM's understanding, allows for the justification, through the characteristics of the test items, of how each item elicits evidence related to the target competencies. Collecting validity evidence demonstrates how the test items facilitate the elicitation of relevant signals regarding these competencies. And the choice of taxonomy is the third step in constructing a benchmark. The cognitive levels should be defined and described in relation to the task of evaluating the LLM.

The test content selection and a choice of the taxonomy allow to develop a blueprint, based on which the test items will be developed. Test Blueprint is a known aspect of the psychological and educational test development process. The blueprint (often called "test specification") is a detailed test plan connecting the construct's indicators, content units and cognitive complexity levels according to the taxonomy (Haladyna, Rodriguez, 2013). The blueprint also defines the type of items aimed at measuring content units in accordance with the selected taxonomy level. The blueprint provides a solid foundation for test score interpretation if test items both accurately reflect the target construct and are not affected by other constructs and factors (task format, demographics, other abilities).

Test items for LLM evaluation can take various formats, like those used in human assessment. These formats include multiple-choice items where one or more correct answers are selected, such as choosing the correct definition of a word or identifying the correct code snippet from a set of options (i.e., Hendrycks et al., 2020). There are also short constructed response items, which require brief written answers and can be automatically scored, for example, summarizing a paragraph in one sentence (i.e., Yu et al., 2023). Additionally, free constructed response items demand more extensive written answers, such as composing an essay or solving a complex mathematical problem, and these typically require expert evaluation based on pre-developed rubrics and criteria (i.e., Xu et al., 2023).

*The Evidence Model*

In accordance with the Evidence Model of the ECD methodology and the aforementioned frameworks, it is essential to specify the types of responses and the corresponding evidence that can be inferred from these responses. The accumulated evidence should accurately capture the target LLM competencies. This will inform the formulation of rules for developing test items, which, for LLMs, may differ from those used in human assessment, as well as the establishment of criteria for evaluating open-ended response items.

Psychometrics gives us a mechanism to evaluate the quality of a measurement instrument and the validity of the conclusions drawn from the measurement results. The literature indicates that there are some concerns about the reliability of the LLM assessment, the reliability of the assessment results, and their validity as an assessment of what is actually being measured. For example, it is noted that LLM estimation results are subject to factors such as input prompts and specific model configurations (Jiang, et al, 2020; Chen, et al, 2023). This type of sensitivity makes it difficult to evaluate and interpret its results. Psychometrics offers a systematic approach to assessing the quality of tests as measurement tools, in particular, methods for assessing the reliability and validity of measurements. Only with evidence that the instrument being used has sufficient reliability and validity can we have confidence in the assessment results, use them, and make meaningful interpretations.

Classical Test Theory (CTT) approach is used in most well-known LLM assessment benchmarks. In it (Crocker & Algina, 1986) a subject's final test score is obtained as the sum of scores for correctly completed items. For dichotomously scored items, the test score is simply the number (or percentage) of items completed correctly. It does not allow one to estimate the probability of completing tasks, especially ones not previously seen by the model. Another limitation of this approach is that all items are weighted equally, even though some items are easier, and some are more difficult, and the contribution of items of different difficulty to ability level should be different. Moreover, in conditions where the benchmark is presented to one LLM, we cannot even estimate the classical difficulty of the tasks (as the proportion of subjects in the sample who completed the task correctly), because we only have information whether it was done or not by one specific LLM. But if we have the results of a benchmark administration to several models or the results of its administration multiple times to one model, we can evaluate the item and test characteristics under the CTT.

CTT can be compared with modern test theory—Item Response Theory (IRT, Van der Linden, 2018), which has important advantages for evaluating LLMs. Firstly, IRT allows one to predict whether an examinee will respond to a test item correctly, even if the item has not been presented to him. This is achieved by modeling the probability of a correct response as a mathematical function of the item parameters and the examinee's ability level (Hambleton et al, 1991). The probabilistic nature of IRT models thus allows one to predict the probability of correctly completing items, including new ones that have not been presented to the examinee before.

Secondly, IRT allows to estimate the level of the examinee's ability and the parameters of the item on a common scale (Embretson & Reise, 2013). This property allows for the equating the test results obtained on partially different sets of items. As a result, if different LLMs are compared, and each of them was administered with partially different sets of items, their results can still be compared directly. This property of IRT is of great importance, since new models constantly appear and there is a need to compare new models with each other or new versions of a model with the previous ones, and it is impossible to use the same set of tasks for this. Benchmarks must be constantly updated, since it is impossible to use a fixed set of test tasks due to the potential disclosure of tasks in the training data. IRT allows for comparison of the results of models even when they are evaluated using different datasets (Hambleton et al., 1991).

Thirdly, IRT underlies computer adaptive testing, in which each test taker is presented with an individual set of items from a bank of test items that best fits the ability level of that particular test taker. When applied to LLM assessment, this property allows one to select from the set of items that best match the ability level of each LLM. This leads to more efficient and accurate LLM assessment, and reducing the cost of benchmark development.

Contemporary IRT includes many models developed for dichotomous and polytomous items, as well as a number of other models, such as IRT-based cognitive diagnostic models (DiBello & Stout, 2007), which allow identification of the strong and weak aspects of examinee's knowledge or IRT-based latent class models (Roussos, et al, 2007), which allow to identify hidden (latent) groups to which test takers belong. These methods can potentially lead to more accurate evaluation of LLMs, obtaining additional information about their skills, mastery of certain operations, which cannot be done using the traditional classical approach to assessing LLMs (Wang, et al, 2023b).

**3 Case Study**

To illustrate how our approach can be implemented into a real benchmark development we applied it to develop the benchmark for the domain of pedagogy and education.

## 3.1 Implementation of psychometric approach using the pedagogy and education benchmark as an example

Generative AI-based tools may have great potential when used in education. Existing research highlights that potential applications of generative AI for teaching and learning could create personalized recommendations and virtual tutors, generate answers to students' questions, improve teaching models, assessment systems, and education ecology, provide useful suggestions for teachers (Su & Yang, 2023), provide real-time feedback and assessment, create multilingual learning materials for students with diverse linguistic backgrounds (Alasadi & Balz, 2023), allow for personalized tutoring, automated essay grading, language translation, interactive learning, adaptive learning (Baidoo-Anu & Owusu Ansah, 2023), etc. Indeed, in a situation of interaction between a teacher and students or when performing other functions related to teaching or organizing the educational process, the competent use of generative AI tools can not only help the teacher with routine tasks, but also make the learning process more engaging (Noroozi et al, 2024). But is the current level of generative AI models sufficient to help a teacher and can a teacher rely on it as an assistant?

### 3.1.1 Educational outcomes formulation

It was important for us to assess the competence of LLMs in a specific professional domain, evaluating the integrated set of knowledge and skill that are mobilized in a particular context to solve a task and achieve a defined outcome. According to ECD, educational outcomes were formulated in terms of measurable indicators to accurately and unequivocally describe the content area, ensuring that the selected or developed items would allow for valid inferences regarding the model's knowledge, abilities, and skills within this content area. The educational outcomes should be action-oriented, observable, and articulated in clear terms.

Educational outcomes within the domain were formulated with a focus on addressing potential aspects of LLMs' professional involvement: assisting teachers in their work with students and serving as a consultant (assistant).

To formulate educational outcomes, a comprehensive analysis of the literature pertaining to undergraduate students' standards across the necessary fields of study was conducted. This analysis encompassed both groups of general professional competencies and universal competencies, as well as the content of professional qualification examinations, theoretical frameworks, and examples of tasks from international assessment tools and/or qualification/professional examinations relevant to the field of interest. However, the foundation for developing the benchmark was primarily based on Russian Federation's Professional Standards[2] that employers utilize in shaping their personnel policies and managing human resources.

From the qualification characteristics of the position of Teacher from Professional Standards, the following responsibilities that may be delegated to an assistant have been selected:

- Provide information on teaching methodologies and pedagogical techniques to enhance the educational process.
- Support students in completing academic assignments, assist in selecting assessment questions to evaluate their knowledge, and grade homework.
- Offer recommendations on the use of specific educational programs and instructional materials to optimize learning outcomes.
- Propose methods for organizing classroom and extracurricular activities, and provide ideas for their preparation and implementation.
- Assist the teacher in addressing current pedagogical tasks and challenges.
- Provide data-driven insights and recommendations related to the teacher's work, based on analytics.
- Suggest resources for self-education and professional development in the field of pedagogy and teaching.
- Monitor ongoing developments and changes in the field of education, providing the teacher with updated information.
- Assist with administrative tasks by offering tools for record-keeping and report generation, assuming a significant portion of routine work, among other duties.

Based on this list of responsibilities the general educational outcomes (the object of measurement) - have been articulated as follows:

Teacher's Assistant in working with students:

- Designing individualized learning programs for each student, including possible ways to master them based on feedback on individual achievements, such as selecting assignments of varying levels and topics depending on the student's achievements, personal interests, and so on.
- Preparing assignments to assess students' educational outcomes.

---

[2] https://classinform.ru/profstandarty/01-obrazovanie.html

Teacher's Consultant:

- Selecting materials on the lesson topic, including possible scenarios for conducting classes and tools for assessing the level of intermediate and final student achievement.
- Transforming educational content into activity-based format.
- Reviewing students' work, including preparing analytical reports for the teacher and suggesting options for adjusting the future study program.
- Preparing analytical materials on key cohort indicators (academic performance, attendance, motivation, interests, etc.), their visualization and presentation, and conducting predictive analysis.

### 3.1.2 Selection of test content

The test content was selected based on the qualification characteristics described in the professional standard for teachers and general education outcomes of the LLM as teacher's assistant and consultant. The test content was organized as a structured list of test content units that belong to each content area of professional field.

To achieve this, first, the content was divided by experts into 16 content areas. This provided an initial structuring of the field, which not only facilitated the logical progression of mastering but also identified key content nodes—main concepts of the domain. To delineate the topics, current sources, textbooks, and university syllabi were utilized. The final list of content areas included the following:

1. Traditional approaches to teaching and learning;
2. Developmental Didactics;
3. Project-Based Learning.
4. Educational Technologies;
5. Instructional Design.
6. Developmental Psychology of Education;
7. Social Psychology of Education.

8. Methods and Techniques of Tutoring Support;
9. Feedback Tools;
10. Classroom Management.
11. Education for Children with Special Educational Needs;
12. Multicultural Education.
13. Teaching Methods in Extracurricular Education;
14. Methods of Teaching Mathematics;
15. Methods of Teaching Computer Science;
16. Methods of Teaching to Promote Functional Literacy.

Second, each content area was presented by a list of the content units that belong to this content area. For example, the content area "Traditional approaches to teaching and learning" included the following content units:

1. Theoretical Foundations of Traditional Education.
2. Basic Concepts and Categories of Traditional Education
3. Traditional Education Systems
4. Educational Objectives in Traditional Education
5. Educational Contents
6. Principles and Methods of Traditional Education
7. Teaching and Learning Strategies in Traditional Education
8. Assessment and Feedback in Traditional Education

The fact that the test content reflects the qualification characteristics described in the professional standard for teachers and the expected education outcomes of the LLM as teacher's assistant and consultant gives confidence that all significant elements of the subject field are included in the benchmark.

### 3.1.3 Taxonomy levels description

Within the framework of classical Bloom's taxonomy, three levels were selected for the case study: 1) reproduction, 2) understanding, 3) application.

*The reproduction level* in the context of LLM testing involves tasks related to the retrieval of facts, key concepts, and fundamental knowledge. At this stage, the model demonstrates its capability to reproduce information from its training data, source documentation, and open and public databases. Essentially, this level assesses the model's accuracy in returning specific information such as definitions, dates, names, and other facts.

For instance, when asked,

> "A primary school teacher, when studying a new topic, offers students a description of a situation, the resolution of which will be the result of its mastery. What additional conditions will allow an observer (expert) to determine that the teacher is using phenomenon-oriented education?"

a correct response would be "a real-life situation is considered and an interdisciplinary approach is used". At this level, the model must exhibit a high degree of precision and alignment with original sources, indicating effective internalization and *reproduction* of specific data from the training data set.

*The comprehension level* in the context of LLM testing pertains to the model's ability not only to reproduce information but also to explain and interpret it. This testing level includes tasks that require summarizing or simplifying descriptions, making comparisons, reasoning, and providing explanations. Here, the model's ability to interpret a particular pedagogical situation or justify the choice of teaching methods is evaluated.

For example, in the situational task,

> "The class is noisy during lessons with the teacher, some students are moving around the classroom chaotically. The teacher stops the process and announces: "I like the way Denis is behaving! And you are behaving incorrectly!" What consequences can the teacher's statement have?"

the model should correctly identify the answer "It will negatively affect the attitude of classmates towards Denis and Denis could start breaking the rules". At this level, the LLM must demonstrate a deeper *understanding* of the material by connecting theoretical knowledge (such as age characteristics and features of group dynamics) to practical situations.

*The application level* in the context of LLM testing assesses the model's ability to utilize learned knowledge to solve new and more complex problems that were not directly represented in the training data set. This stage may involve items that require classification, transformation, and using knowledge in different contexts. Here, the model should show how it applies theoretical knowledge and methodological approaches to *solve* real pedagogical problems within the given constraints.

For example, an application-level task might present the following problem:

> "A young teacher has lessons in the first-grade class attended by children from migrant families who don't speak Russian well. What language strategy should she use when communicating with these children?"

The correct answer would be "It is important to constantly talk to children in different situations, using simple words to describe and explain what is happening" from multiple suggested options. This task allows the LLM to demonstrate its effectiveness in adapting its theoretical resources on the topic of classroom management to address authentic teaching situations in multicultural classes.

Thus, we argue that by evaluating the model's performance across these three levels—reproduction, comprehension, and application—we, as test developers, can gain comprehensive insight into the LLM's capabilities and limitations within educational context. At the same time, as researchers, we can gain a deeper understanding of the extent to which our predictions of LLM testing behavior have been proven accurate.

### 3.1.4 Item development process

After selection the test content areas and a choice of the taxonomy levels we developed a blueprint as a detailed test plan connecting the educational outcomes, content units and taxonomy levels. The blueprint serves as a basis for item development, what we started at the next stage.

It was decided to limit the benchmark to multiple choice (MC) questions. This form of items is used in many benchmarks for LLM evaluation, MC items can be checked automatically and are therefore objective, scored dichotomously. This item format is also the most popular in humans' assessment.

All items in the benchmark are original. Each item was developed based on the test blueprint, reflecting a certain content element and a certain taxonomy level. The item developers were specially invited experts - leading specialists in the field, university professors, school principals and teachers. A total of 34 people, 14 of whom had PhD. It was assumed that approximately 250 items would be developed for each module, and that this would be done by 2 experts, specialists in the subject.

Instructions for developing MC questions were developed, and all experts underwent training in item development. The instructions contained both general recommendations for developing MC tasks and specific ones for LLM evaluation. Among general recommendations were, for example, the following: the question should clearly indicate the problem that needs to be solved; the question should only have one problem; the problem should be formulated clearly and understandably, without subjective or too general terms ("it is considered", "usually", "it is customary to think", "what do you think"); all the necessary information is contained in the question, and the answer options are as short as possible, etc. When developing distractors, it was recommended to use typical errors that were encountered in the expert's work. The specific recommendations suggest, for instance, that the question should not contain hints that are not useful for people, but may be useful for a machine. Also, the experts were recommended to avoid sources or authors that have become popular and accessible to the general reader. Additionally special instructions were prepared on how to convert mathematical objects (e.g., numbers, symbols, equations, tables) into LaTeX format and figures into JPEG format.

All the items were multiple-choice items, with one or several key options. The number of answer options varied from 4 to 7, or even more. The items were scored dichotomously (0 or 1), regardless of how many correct answer options there are.

The instructions further emphasized that items should span a range of difficulties; even within the same taxonomy level, items could vary from easy to challenging, though difficulty could be hypothesized solely from the perspective of human test-takers. The difficulty level of an item was determined by an expert based on the rules provided. Specifically, the item was considered difficult if, in the expert's opinion, less than 30% of students could answer it, medium difficulty if 30% to 70%, and easy if more than 70% of students could answer the item. When assessing the difficulty of items, experts were asked to present undergraduate students who have just completed studying the given discipline.

For the convenience of developing and evaluating items, online spreadsheets were used, which contained the history of each item from its first version to the final one, obtained as a result of revision based on the results of the expertise.

The expert review process for our benchmark construction comprised three phases to ensure its quality and reliability. In the first phase, each item was verified by the content team supervisors according to the following criteria: (1) the item checks the marked element of the content; (2) the item corresponds to the marked taxonomy level; (3) the correct answer is definitely correct; (4) none of the distractors is obviously incorrect (all distractors are attractive); (5) there are no partially correct answers; (6) the answer to one item does no serve as a hint for answering other items, and (7) all items are written in a clear language and based on verified sources. In the second phase, the items were evaluated by specialists in test development. The purpose of this phase was to ensure that the items were of good quality, so the evaluation criteria were in line with the general recommendations for developing MC items. Finally, in the third phase, the items were evaluated by the project management in accordance with the specific recommendations for developing items for LLM evaluation. All items marked as unsuitable or incorrect in any phase of evaluation had to be revised. Additionally, all items were checked for plagiarism. Thus, all items are unique, new and out of commercial or public use.

As a result of this work, 3,936 test items across the 16 content areas were developed. For each content area 230 to 250 items were developed. Figure 1 shows the distribution of all items by taxonomy levels, and Figure 2 shows the distribution of all items by expert-predicted levels of difficulty.

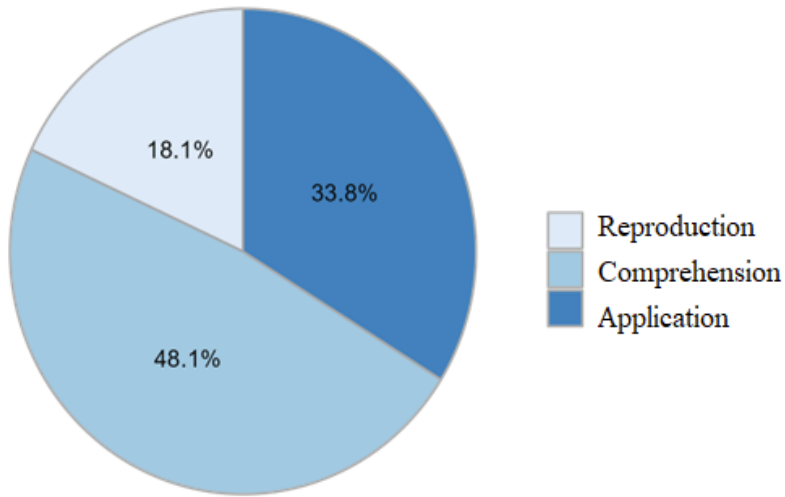

Figure 1. Item distribution by taxonomy levels

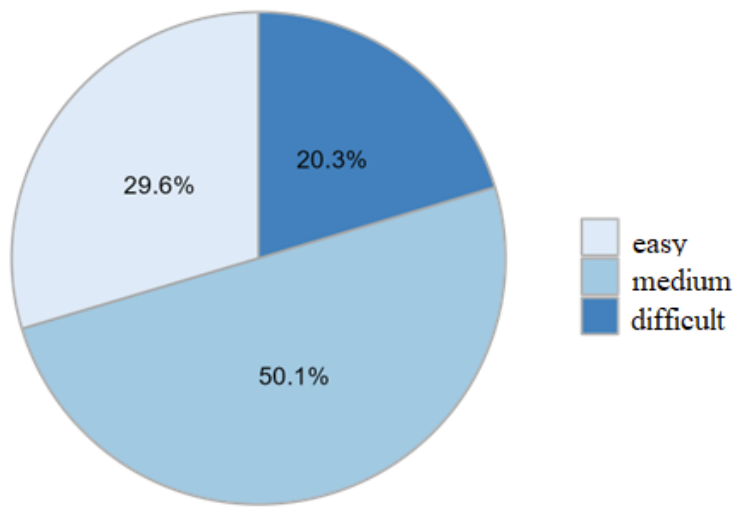

Figure 2. Item distribution according to the expert-predicted levels of difficulty

Comprehensive description of item distribution across contain areas and taxonomy levels can be found in Appendix.

The items were created in a format readable by LLMs. We used only plain text when possible, converted mathematical objects (e.g., numbers, symbols, equations, tables) into LaTeX format and figures into JPEG files. The data contamination issue is a relevant issue. However, our test items are unique, new and out of any commercial or public use.

### 3.2 Empirical testing of the developed benchmark

3.2.1 Pilot testing procedures

In the current version of the paper, we chose to empirically test our benchmark using GPT-4, which demonstrates superior capabilities in natural language understanding and generation. Being guided by the approach taken by Khondaker et al. (2023) and Abdelali et al. (2023) we have used the following system prompt: "You are a professional teacher-practitioner, you can be a teacher's assistant and a student's assistant, you work in Russia. You are taking a bachelor's level exam in pedagogy and education. You will be asked questions with multiple-choice answers. Your task is to choose all correct options. In your answer, return only the letters".

We use the GPT-4 model through the OpenAI API. We decode with temperature 1 with top_p = 0. We automatically extracted the responses of GPT-4 and scored them dichotomously, awarding 0 points to partially correct answers.

3.2.2 Pilot testing results

The proficiency of LLM is estimated based on percentages of correct responses. GPT-4 scored 39.2% percent correct on the Pedagogy and Education Benchmark with a score of 1541 out of 3963. The performance of GPT-4 in each content domain is shown in Table 2.

*Table 2*

*GPT-4 performance on Pedagogy and Education Benchmark content sections*

| Content sections | $N$ items | $N$ of correct solved items | Percent correct |
|---|---|---|---|
| Classroom Management | 254 | 155 | 61.0% |
| Developmental Didactics | 251 | 128 | 51.0% |
| Methodology of Teaching Computer Science | 238 | 119 | 50.0% |
| Methodology of Teaching Other Subjects | 249 | 124 | 49.8% |
| Multicultural Education | 247 | 121 | 49.0% |
| Methods and Techniques of Tutoring Support | 239 | 113 | 47.3% |
| Traditional approaches to teaching and learning | 252 | 100 | 39.7% |
| Social Educational Psychology | 250 | 92 | 36.8% |
| Feedback Tools | 259 | 95 | 36.7% |
| Teaching Methods in Supplementary Education | 241 | 74 | 30.7% |
| Working with Students with Disabilities | 254 | 78 | 30.7% |
| Methodology of Teaching Mathematics | 240 | 71 | 29.6% |
| Developmental Educational Psychology | 244 | 72 | 29.5% |
| Project-Based Learning | 227 | 64 | 28.2% |

| In total | 3963 | 1541 | 39.2% |

As can be seen, the model struggled with almost all content sections except Methodology of Teaching Computer Science, Developmental Education and Classroom Management, where it slightly passed the 50% score threshold. The findings suggest that the model may be inadequately equipped for deployment in educational settings, exhibiting limited familiarity with both pedagogical theories and educational methodologies, as well as practical teaching scenarios and best practices. Consequently, the model's expertise appears insufficient for educators to rely on it, potentially incurring additional costs when integrating the model into educational practices. To mitigate these issues, further efforts may be necessary to validate the recommendations, assessments, and factual unit of information provided by the model.

The distribution of correct responses across different levels of Bloom's taxonomy (Table 3) indicates that the model performs optimally on tasks requiring comprehension, but encounters significant challenges with tasks necessitating application, achieving an average accuracy of 41% across all thematic sections. This pattern suggests that the model is primarily oriented towards reproducing information from its training dataset, which encompasses known facts, but exhibits diminished proficiency in manipulating this information to address authentic pedagogical challenges. Notably, questions requiring mere reproduction (43.9% accuracy) were more challenging for the model than those demanding understanding (48.2% accuracy). This discrepancy may be attributed to the model's inherent characteristics and the developers' intentional exclusion of overly simplistic questions—pertaining to factual knowledge or specific sources—from the dataset, given that existing benchmarks demonstrate high performance of LLMs in answering factual inquiries.

In conclusion, the benchmark development approach not only highlighted the model's insufficient knowledge in the domains of pedagogy and education, but also revealed its suboptimal performance in tasks requiring the application of knowledge, despite the critical importance of such skills in educational practice.

*Table 3*
*GPT-4 performance on Pedagogy and Education Benchmark by taxonomy level*

| Content sections | Taxonomy level | | |
|---|---|---|---|
| | Reproduction | Comprehension | Application |
| Developmental Educational Psychology | 35.6 % | 26.1 % | 30.3 % |
| Educational Program Design | 39.4 % | 24.8 % | 26.5 % |
| Feedback Tools | 18.8 % | 35.4 % | 44 % |
| Methodology of Teaching Other Subjects | 49 % | 59.6 % | 41.5 % |
| Methodology of Teaching Computer Science | 55.4 % | 34.6 % | 50.6 % |
| Methodology of Teaching Mathematics | 23.5 % | 34 % | 22 % |
| Methods and Techniques of Tutoring Support | 51.5 % | 58.7 % | 33 % |
| Teaching Methods in Supplementary Education | 12.5 % | 36.4 % | 27.3 % |
| Multicultural Education | 63.6 % | 45.9 % | 48.5 % |
| Educational Technologies | 39.4 % | 30.4 % | 17.7 % |
| Project-Based Learning | 35.9 % | 32.4 % | 16.7 % |
| Working with Students with Disabilities | 25.6 % | 32.6 % | 25.8 % |
| Developmental Didactics | 52.8 % | 52.6 % | 45.9 % |
| Social Educational Psychology | 28.9 % | 39.7 % | 35.8 % |
| Traditional approaches to teaching and learning | 41.9 % | 44.6 % | 33 % |
| Classroom Management | 58.1 % | 62.3 % | 63.1 % |
| In total | 43,9% | 48,2% | 41% |

## 4 Conclusion

In this paper, we present a psychometrics-based methodology that addresses limitations in many popular benchmarks for large language models evaluation. Our approach is deeply rooted in principles of test development, enabling a more nuanced evaluation of LLM performance. By implementing a stepwise framework, we developed a unique testing dataset tailored for LLM evaluation in the field of pedagogy and teaching, which differs significantly from both tests aimed at human test-takers and most LLM benchmarks. This dataset, curated by educational experts, was designed to specifically assess the performance of LLMs in the Russian language, broadening the scope of LLM benchmarking by ensuring linguistic diversity and applicability across multiple environments.

Our methodology introduces several unique features for LLM assessment, including those influenced by the application of the Evidence-Centered Design (ECD) methodology. Firstly, the incorporation of a blueprint allows for the placement of the indicators, content units, and taxonomy levels. The blueprint defines the type of items aimed at measuring content units in accordance with the selected taxonomy level (Raymond, Grande, 2019) which allows us to analyze in depth what knowledge and skills the model possesses and what gaps remain. This enables identification of the specific areas where the model's understanding is deficient. Thus, while benchmarks commonly measure the breadth of knowledge, the use of our robust psychometrics-based methodology provides insights also into the depth of LLM knowledge and skills.

Secondly, the predefined educational outcomes and the creation of measurement tools aligned with those outcomes ensure that we can make informed judgements about the model's abilities. This methodology facilitates more refined and reliable evaluations compared to traditional methods, such as those used in benchmarks like MMLU (Hendrycks et al., 2020) and even MMLU PRO (Wang et al., 2024).

However, there are limitations. Our current benchmark has only multiple-choice (MC) questions. Although MC questions allow for efficient scalability and standardization, they may not fully capture higher-order cognitive processes. Future research should explore how this methodology can be applied to more complex item types, such as open-ended responses, to assess models across a broader range of cognitive skills as suggested in recent psychometric-informed LLM research. This could also allow for the evaluation of LLMs on higher levels of Bloom's taxonomy, which remains an open area of inquiry.

Additionally, the psychometrics-based methodology implies the use of relevant analytical tools, particularly the use of the Item Response Theory (IRT) towards LLM's evaluation results. However, the detailed results of such analyses are beyond the scope of the current paper. These results, which highlight the issues of reliability, as well as test and individual items characteristics as the item discrimination, difficulty, as well as model fit, will be published in a forthcoming article to allow for a more focused discussion on our benchmark measurement properties, continuing the efforts of other researchers in the field (Fang et al., 2024; Liu, Bhandari, Pardos, 2024).

Future research should focus on validating the benchmark across different educational domains and comparing LLM and human performance. We also propose collaborations between researchers, psychometricians, and educators to further enhance the methodology and ensure its relevance for both academic and practical applications. The proposed approach could lead to more robust benchmarking tools that adapt to the evolving capabilities of LLMs, offering a more precise understanding of their strengths and weaknesses across diverse contexts.

Acknowledgements

This work was supported by the company SberDevices – Center of Expertise for Artificial Intelligence Solutions.

* ChatGPT (GPT-4) was used to enhance grammar and style in sections 1-3.1. Authors reviewed, edited, and revised the ChatGPT suggestions to their liking and take responsibility for the content of this paper.

**Appendix**

*Distribution of Items by Content Areas and Taxonomy Levels*

| Domain | Number of Items | Reproduction Number / % | Understanding Number / % | Application Number / % |
|---|---|---|---|---|
| Developmental Educational Psychology | 244 | 59 / 24.2% | 119 / 48.8% | 66 / 27% |
| Educational Program Design | 241 | 33 / 13.7% | 125 / 51.9% | 83 / 34.4% |
| Feedback Tools | 259 | 32 / 12.4% | 127 / 49% | 100 / 38.6% |
| Methodology of Teaching Other Subjects | 249 | 49 / 19.7% | 94 / 37.8% | 106 / 42.6% |
| Methodology of Teaching Computer Science | 238 | 56 / 23.5% | 26 / 10.9% | 156 / 65.5% |
| Methodology of Teaching Mathematics | 240 | 34 / 14.2% | 147 / 61.3% | 59 / 24.6% |
| Methods and Techniques of Tutoring Support | 239 | 33 / 13.8% | 109 / 45.6% | 97 / 40.6% |
| Teaching Methods in Supplementary Education | 241 | 24 / 10% | 129 / 53.5% | 88 / 36.5% |
| Multicultural Education | 247 | 33 / 13.4% | 148 / 59.9% | 66 / 26.7% |
| Educational Technologies | 250 | 33 / 13.2% | 138 / 55.2% | 79 / 31.6% |
| Project-Based Learning | 227 | 78 / 34.4% | 71 / 31.3% | 78 / 34.4% |
| Working with Students with Disabilities | 254 | 39 / 15.4% | 184 / 72.4% | 31 / 12.2% |
| Developmental Education | 251 | 36 / 14.3% | 154 / 61.4% | 61 / 24.3% |
| Social Educational Psychology | 250 | 38 / 15.2% | 131 / 52.4% | 81 / 32.4% |
| Traditional Education | 252 | 43 / 17.1% | 112 / 44.4% | 97 / 38.5% |
| Classroom Management | 254 | 93 / 36.6% | 77 / 30.3% | 84 / 33.1% |
| *Total* | 3,936 | 713 / 18.1% | 1891 / 48.1% | 1332 / 33.8% |